\documentclass[preprint,3p,twocolumn]{elsarticle}

\usepackage{lineno,hyperref}
\usepackage{amsmath,amssymb,amsfonts}
\usepackage{algorithmic}
\usepackage{graphicx}
\usepackage{textcomp}
\usepackage{soul}

\usepackage{booktabs}
\usepackage[table]{xcolor}
\usepackage{tabularx}
\newcolumntype{Y}{>{\centering\arraybackslash}X} 
\usepackage{tikz}
\usepackage{import}
\usepackage{siunitx}
\usepackage{csvsimple}
\usepackage{subcaption}
\usepackage{csquotes}

\definecolor{sampleBlue}{rgb}{0.12, 0.47, 0.71}	
\definecolor{darkPurple}{rgb}{0.27, 0, 0.33}		
\definecolor{darkGreen}{rgb}{0.13, 0.56, 0.55}		
\definecolor{darkYellow}{rgb}{0.99, 0.90, 0.15}	
\definecolor{lightPurple}{rgb}{0.92, 0.90, 0.93}	
\definecolor{lightGreen}{rgb}{0.91, 0.95, 0.95}	
\definecolor{lightYellow}{rgb}{1, 0.99, 0.91}		

\modulolinenumbers[5]

\journal{Advanced Engineering Informatics Journal}

\begin{document}

\begin{frontmatter}

\title{Concept Identification for Complex Engineering Datasets}

\author{Felix~Lanfermann\corref{mycorrespondingauthor}}
\cortext[mycorrespondingauthor]{Corresponding author (felix.lanfermann@honda-ri.de)}

\author{Sebastian~Schmitt}

\address{Honda Research Institute Europe GmbH, Carl-Legien-Strasse 30, D-63073 Offenbach/Main, Germany}

\begin{abstract}
Finding meaningful concepts in engineering application datasets which allow for a sensible grouping of designs is very helpful in many contexts. 
It allows for determining different groups of designs with similar properties and provides useful knowledge in the engineering decision making process. 
Also, it opens the route for further refinements of specific design candidates which exhibit certain characteristic features.
In this work, an approach to define meaningful and consistent concepts in an existing engineering dataset is presented.
The designs in the dataset are characterized by a multitude of features such as design parameters, geometrical properties or performance values of the design for various boundary conditions.  
In the proposed approach the complete feature set is partitioned into several subsets called description spaces. 
The definition of the concepts respects this partitioning which leads to several desired properties of the identified concepts.
This cannot be achieved with state-of-the-art clustering or concept identification approaches.
A novel concept quality measure is proposed, which provides an objective value for a given definition of concepts in a dataset. 
The usefulness of the measure is demonstrated by considering a realistic engineering dataset consisting of about 2500 airfoil profiles, for which the performance values (lift and drag) for three different operating conditions were obtained by a computational fluid dynamics simulation. 
A numerical optimization procedure is employed, which maximizes the concept quality measure and finds meaningful concepts for different setups of the description spaces, while also incorporating user preference.
It is demonstrated how these concepts can be used to select archetypal representatives of the dataset which exhibit characteristic features of each concept.
\end{abstract}


\end{frontmatter}



\section{Introduction}
\label{sec:introduction}

In an engineering design process, different candidate designs are created.
They are further analyzed in order to identify desirable solutions and design concepts.
A design concept incorporates different designs that are similar with respect to their features as well as the associated quality criteria.
Design concepts provide valuable insight into the design problem itself during the engineering decision making process. They allow insights on the distribution of groups of solutions, their corresponding quality criteria, and general proximity relations among solutions~\cite{Rosch1975}.
They also enable the engineer to derive archetypal samples, which represent design types and features  which are of interest to the designer in a specific context.
Representatives can, for example, be drawn such that they describe an archetypal configuration which represents the characteristic  features of that concept.
These representatives can be used as prototypes for further design stages, such as refinement of the initial design. 
They may also serve as starting points for subsequent optimization studies under changed boundary conditions or case-based reasoning approaches~\cite{kolodner2014}.
Since prototypes from different concepts can represent different parts of the search space, they can also be used to realize improvement potential in multiple directions in the search space~\cite{leImp2013}.

In the context of engineering tasks, two types of feature sets are traditionally considered to be the most relevant. 
For one, the set of parameters specifying the design, typically called design parameters, and for another, the set of performance criteria extracted from the designs~\cite{Graening2014, Lanfermann2020}. 
Consequently, the definition of concepts operate on those two feature types. 
The problem of concept identification amounts to finding various groups of designs, where solutions with similar design parameters and similar performance values are grouped into one concept~\cite{Adajian2005,Laurence1999}.

However, additional types of feature sets naturally arise during the complete engineering design process. 
To create robust solutions, candidate designs are usually evaluated for different operating conditions, each one giving a separate set of performance values.
Moreover, additional geometric features are typically calculated from the designs in order to check certain design requirements.
Therefore, it is necessary to extend the definition of concepts to consider all those different types of feature sets. 
And designs need to be grouped in such a way that they show similar values for all different types of features.

Identifying design concepts for a given engineering problem is typically not unique and multiple distinct concepts can be defined.  
Therefore, it might often be desirable to integrate user-preference into the process, as this serves as an anchor for some desired properties of the concepts and removes ambiguities in the concept definitions.
An intuitive way to express user preference is by choosing few candidate solutions which are of interest to the engineer, i.e.\ which share desired properties~\cite{wangPreference2017}.
The concept identification process should then be able to define concepts around these preferred solutions. 

In summary, two separate research questions need to be addressed:
How can design concepts be identified if multiple types of feature sets are to be taken into consideration?
How can user-preference be integrated into the concept identification process?

To address these questions, a unified and general approach for defining concepts for engineering problems and datasets is presented in this work.  
The notion of a \emph{description space} is introduced, referring to a subset of the full feature set which characterizes the designs in a specific context. 
Although the parameter space defining the design, as well as the objectives characterizing the performance at specific boundary conditions, constitute natural description spaces, the partitioning of features into description spaces is in principle arbitrary.
They can be chosen by the engineer based on some preference and thus determine the feature type and their correlations, which will allow the discrimination between concepts. 

At the core of the proposed concept identification method is a concept quality measure (CQM).
The CQM assigns a numerical value between zero and one to a proposed set of concepts in a dataset.
The CQM value reflects how good the proposed concepts are, where zero indicates concepts of low quality while one refers to a very consistent and meaningful concept. 
The assessment of concept quality takes concept size, overlap, and consistency across description spaces into account. 

The question whether the identified concepts are meaningful or not, strongly depends on the engineer's expectations and the context in which concepts are searched. 
It is difficult to formulate user expectation in a general and abstract way. 
But, typically, engineers can express their preferences in terms of selected data samples. 
Doing so, they specify which data samples should be in the same or in different concepts, or which feature values are more important than others. 
Therefore, the CQM allows for the inclusion of user preference which is provided by the engineer as special sets of feature vectors.

With this measure at hand, a concept identification approach is proposed.
It is based on a numerical optimization procedure which parametrizes the regions associated with each concept and then maximizes the CQM by adjusting this parametrization.
The approach is thoroughly discussed and its effectiveness is shown at an example of a design dataset obtained from an aerodynamic airfoil optimization process.  
A definition of the key terms used in this work can be obtained from Table~\ref{tab:terms}.

The remainder of the paper is structured as follows: Section~\ref{sec:related_work} summarizes related work and similar methods from the fields of optimization, clustering, and concept identification.
Section~\ref{sec:methods} introduces a metric to objectively evaluate multiple concepts in a dataset based on multiple description spaces and describes how user-preference is addressed by integrating specific samples of interest into the concept identification process.
In section~\ref{sec:experiments}, an instructive example, followed by experiments on an engineering design dataset demonstrate the necessity and applicability of the proposed methodology.
Concluding remarks are given in section~\ref{sec:conclusion}.

\begin{table*}[!ht]
\centering
\caption{Description of key terms}
\begin{tabularx}{0.98\textwidth}{lX}  
\toprule
Term & Description \\
\midrule
design & one candidate solution to the engineering problem at hand\\
feature & a property that is attributed to a design \\
data sample, feature vector & one full set of features associated with one design\\
dataset & a set of designs with their corresponding features vectors \\
description space & the space that is spanned by a group of features\\
concept & a set of data samples that is similar in more than one description space\\
concept quality metric & a numerical quality measure for one set of concepts \\
concept identification & the separation of data samples into concepts by optimizing a concept quality metric\\
\bottomrule
\end{tabularx}
\label{tab:terms}
\end{table*}

\section{Related Work}\label{sec:related_work}

Finding high performing yet different solutions to a technical problem is an exciting endeavor that has been approached in various domains.
It can be viewed from different scientific angles, ranging from the classification perspective of clustering algorithms and diversity generation in \mbox{(multi-/many-)} objective evolutionary optimization to the prototype theory of concepts.
Naturally, the developed methods are tailored towards specific problems and hence not universally applicable.

\subsection{Optimization-Based Approaches}\label{subsec:state_of_art_optimization}

Within the domain of evolutionary computation, there are several methods that specifically target finding high-performing diverse solutions.
Different types of niching~\cite{Mahfoud1995_c}, tabu search~\cite{Glover1993}, and restarting schemes~\cite{Beasley1993} use mostly distance-based measures with respect to either the design parameters or the performance space.

Other approaches define and derive additional properties of the solutions and incorporate them into the development process.
This is, for example, done by relying on novelty alone to select relevant solutions and steer the optimization into the desired direction~\cite{Lehman2011}, by combining novelty and a local competition factor of the considered solutions~\cite{Lehman2011b}, or restarting the optimization based on a novelty-score~\cite{Cuccu2011}.
Furthermore, additional derived measures such as the interestingness~\cite{Graening2010,Reehuis2013} or the curiosity~\cite{Schaul2011} can be considered in addition to performance criteria.

Illumination algorithms~\cite{Mouret2015} and quality diversity algorithms~\cite{Pugh2016}, such as novelty-search with local competetion (NSLC)~\cite{Lehman2011b} and multi-dimensional archive of phenotypic elites (MAP-Elites)~\cite{Mouret2015} explicitly consider one additional description space besides design and performance space.
Where in NSLC, the space of behaviors is considered to judge the novelty of the designs, a (in practice low-dimensional) space of arbitrarily definable features serves as the basis for exploration in MAP-Elites.
The unifying framework of quality and diversity optimization aims at producing \enquote{a large collection of solutions that are both as diverse and high-performing as possible}~\cite{Cully2018}.

These reviewed methods aim at finding independent novel solutions of high quality. 
They are either looking for such solutions based on the typical separation of features into a parameter and an objective space, or by integrating one additional feature space.
Nevertheless, they mainly focus on finding single high-performing solutions and lack the identification of groups of solutions which are similar in a multitude of relevant feature spaces, i.e.\ description spaces.
But the discovery of such groups is essential to the identification of concepts.
It is thus argued that, despite achieving outstanding performances in other application scenarios, the above-listed  optimization-based approaches are not suitable for concept identification.

\subsection{Clustering Approaches}\label{subsec:clustering}

Concept identification may technically be viewed as a special type of clustering problem with  specific constraints on the returned solution. 
Clusters and concepts share the notion that samples with similar features should be associated with each other. 
However, a clear distinction between clusters and concepts is made in this work. 
Clusters represent groupings based on a similarity which considers all features simultaneously.   
In contrast, concepts, as defined in this work, consider a partitioned set of features, i.e.\ multiple disjunct orthogonal projections of the original feature set. 
Within each projected feature space, samples are associated with concepts considering all features of that particular space. 
But a concept has to obey the additional constraint, that the same samples need to be associated with the same concept in all projected feature spaces simultaneously.

Typical clustering methods are not suited to solve the concept identification problem of this work for a number of reasons.
First, it cannot be expected that the complete feature space can be partitioned into concepts, and consequently some design solutions should not be part of any concept.  
This renders the use of classical clustering approaches such as k-means~\cite{MacQueen1967} or k-means++~\cite{Arthur2007} difficult as they always cluster the complete feature space.
But even approaches, that allow for less strict association, such as fuzzy c-means~\cite{Bezdek1981} and Gaussian mixture models~\cite{Rasmussen2000}, are not well suited to handle multiple description spaces.
They identify conjunct clusters of samples based on their feature vector, but they cannot respect the partitioning of the feature vector into description spaces.
It is not possible to identify corresponding groups in several projections of the full feature space simultaneously, where each concept is be represented by a cluster of designs in the space of design parameters, a corresponding cluster in the space of performance values, and additional corresponding clusters of other feature sets. 

State-of-the-art clustering approaches cannot be easily extended to realize such a partitioning of the feature space in a straight forward manner and are thus not suited to approach the concept identification problem~\cite{Lanfermann2020}.
Additionally, the fundamental notion employed by distance- or density-based clustering approaches, that all points within a certain region of the feature space belong to the corresponding concept, cannot be upheld in general due to the possibly incomplete feature vectors. 
Some specific design solutions might be different in certain description spaces, but cannot be distinguishable in another description space due to the incomplete feature set. 
These solutions should not be associated with any concept and thus should not be considered in the clustering approach.
This renders the use of classical approaches, even in projected subspaces, i.e.\ the description spaces, difficult.

Clusters in lower-dimensional projections are identified when subspace clustering~\cite{Parsons2004} is applied on high-dimensional (sparse) data.
However, the method's primary aim is data-driven dimensionality reduction and coping with the sparsity of the data. Therefore, the projections of interest are not predetermined in advance.
But this is a requirement for the concept identification problem considered here, as the projections need to be aligned with the user defined description spaces. 
Finally, subspace clustering might (intentionally) lead to overlapping clusters~\cite{Sim2013}, which the proposed approach explicitly avoids.

Multi-view clustering~\cite{Bickel2004}, focuses on clustering data that can be described by multiple views (multi-view data), where a \enquote{view} can be understood as a group of connected features that describe the data.
These data hence have heterogeneous properties which have a potential connection~\cite{Yang2018} and often contain complementary information that should be exploited~\cite{Yang2018}.
Rather than clustering data based on all features directly, multi-view clustering methods try to maximize a clustering quality metric within each view, while accounting for clustering consistency across different views~\cite{Yang2018}.
For example, a multi-view spherical k-means implementation achieves this by alternating the maximization and expectation step for each view of the data~\cite{Bickel2004}.
A second implementation, particularly tailored toward high-dimensional data, aims at high clustering performances by optimizing a weight factor for each view in the k-means clustering loss-function~\cite{Cai2013}.
Furthermore, multi-view subspace clustering allows for uncovering underlying subspaces in each view while clustering multi-view data~\cite{Gao2015}. 
Multi-view clustering techniques prioritize optimizing the clustering performance by utilizing multiple views.
The concept of a view is rather similar to our concept of a description space. 
However, the consistency within different views is accounted for differently. 
In multi-view clustering, maintaining cluster consistency is factored in as a trade-off relation to the individual cluster quality measures.  
In contrast, consistency is the central requirement for the proposed concept identification approach, which cannot be  guaranteed with multi-view clustering.

\subsection{Concept Identification Approaches}\label{subsec:data_mining_concept_identification}

A number of metrics have been directly developed to evaluate association patterns of data samples by quantifying the relation between groups of samples in different projections of a data set.
In an engineering design framework, these metrics can be used to identify concepts.

A metric that calculates the interestingness and significance (IS) was developed to evaluate the quality of association patterns and overcome counter-intuitive evaluation results~\cite{Tan2000}.
A utility measure was introduced to incorporate an additional indication for concept performance~\cite{Graening2014}.
Both measures, however, lack the ability to consider concept overlap and the total extent of multiple concepts.
They do not evaluate concepts differently if, e.g., two concepts share a majority of the same data samples. This might render the identified concepts insignificant.

To mitigate the shortcomings of the existing measures with respect to concept overlap, a measure was proposed in a previous work to identify and evaluate sets of concepts by explicitly taking their sizes and overlap into account~\cite{Lanfermann2020}.
The proposed measure was developed to identify a set of optimally distributed concepts, but considers only two description spaces.
However, as described above, in many engineering design tasks more than two description spaces naturally arise and the proposed measure cannot be directly applied.

\section{Concept Evaluation based on Multiple Description Spaces and User Preferences}\label{sec:methods}

In the following, first some considerations on the notion of description spaces and its relation to concepts are provided.
Then, a metric to objectively evaluate multiple concepts in a dataset based on multiple description spaces is defined.

\subsection{Description Spaces and Concepts}

An engineering design process is a complex task in which a vast number of different designs are created.
The process needs to be well organized in order for the engineer to be ahead of it. 
Therefore, characteristics of different designs, or groups of designs, should be easily graspable by the engineer. 
This is especially true in the early stages of the process where many fundamentally different variants are explored. 
In this context, defining groups of prototypical designs and concepts is very helpful. 

Typically, the definition of concepts in engineering datsets is based on design parameters and one set of features that describe the performance~\cite{Graening2014, Lanfermann2020,Adajian2005,Laurence1999}. 
A design refers to a solution to a technical problem which can typically be constructed by applying certain rules or processes to adjustable design parameters. 
Performance features, for example production costs and operating efficiency, are associated with each design.

However, apart from these two sets, many additional types of feature sets naturally occur. 
For one, the performance of each design is evaluated for multiple operating conditions, each of which gives a separate set of features. 
For another, geometric features might be calculated from the design, such as local curvatures or thickness values at specific locations of the design, which can be used to specify requirements or to assess manufacturability.  
Additionally, recent approaches utilize machine learning  and geometric deep learning techniques such as point cloud auto-encoders~\cite{Rios2021} to generate low dimensional latent representations.
These characterize the geometric appearance of each design and, therefore naturally qualify as feature sets for each design.
All these feature sets provide different informational aspects of each design and it is hence sensible to separate them logically.
In the approach, the full feature vector is therefore partitioned into these different sets of features. 
Each of those sets denotes a description space, as the characterization of the design with those features of one description space reveals information on the design (with respect to that particular category of features).     
Therefore, it is natural to extend the definition of concepts to consider all those different types of feature sets, i.e.\ descriptions spaces, and group designs in such a way that they show similar values for all different description spaces. 

Another aspect found in practical engineering applications is the fact that the feature vectors associated with the designs typically do not capture all relevant aspects of the design consistently. 
For example, geometric features and design parameters might not capture material properties or manufacturing related issues.
But the latter might have a strong influence on some objectives such as cost or weight.
If not all relevant features are explicitly captured, the feature vectors are only an incomplete description of the design. 
If the designs are incompletely described, a consistent and complete separation of all designs into concepts, is not possible.
As a consequence, in practical applications, some designs cannot be associated with any concept.
Designs which differ in some types of features, e.g.\ the manufacturing objective, might exhibit very similar values for all other feature types, since the features with the discriminative power are missing from the description.
This implies that a sensible concept identification process must be able to deal with such ambiguity and still deliver concepts which make sense to the engineer.

As described earlier, some sets of features such as the design parameters share a semantic context and can thus be naturally considered as one description space. 
However, the partitioning of the features into description spaces is in principle arbitrary and the engineer can choose a partitioning which is most useful for a given context. 
For example, assigning two specific features to different description spaces allows for concepts which show strong correlations between the values for those two features.
Additionally, not every feature needs to be incorporated in a description space. 
This reflects the freedom to choose which features should be relevant for the concepts and which features can be discarded during the concept identification process.

\subsection{Concept Quality Measure and Concept Identification}

In contrast to existing metrics, the proposed metric allows to assess more than two feature sets.
Each feature set constitutes a description space and can contain any number of features.
The proposed approach determines the best partitioning of the entire dataset into concepts and identifies only those data samples which consistently adhere to these concepts. 
Thereby, the problem of the incomplete feature vectors is addressed and consistent concepts are derived. 
Moreover, user-preference can be integrated by providing specific samples of interest into the concept identification process.

A set of $N_\mathcal{D}$ data samples
\begin{equation}\label{eq:dataset}
    \mathcal{D} = \left\{x_i, i = 1, ...,  N_\mathcal{D} \right\}
\end{equation}
is considered, where each data sample is characterized by $N_F$ features $f_j$, i.e.\ $x_i=(f_1,\dots,f_{N_F})^T \in \mathbb{R}^{N_F}$.
Features are then partitioned into separate groups, each forms a description space. 
Without loss of generality, the features can be ordered according to the description spaces and each feature vector can then be written as 
\begin{align}
    x=(x^1,\dots,x^{N_{DS}}, x^\Delta)^T\,,
\end{align}
where $x^k$ ($k=1,\dots,N_{DS}$) is the projection of the feature vector into description space $k$. 
Notice, that each description space can have a different dimension, depending on how many features are associated with it.
The set of features which are not associated with any description space is also included and denoted by $x^\Delta$. The term $x^\Delta$ can also be empty.

A notion of a concept is employed which implies that different samples associated with the same concept are similar to each other in each description space. 
Candidates which could be associated with a concept $\alpha$ can then be selected based on their similarity of the features in each description space.
An intuitive way to do that is to define a simply connected region in each description space and regard all data samples inside this region as potential candidates for the corresponding concept.  
Such a set of data samples inside this region which are potential candidates for concept $\alpha$ in description space $k$ is denoted by $C_{\alpha,  k}$, where $\alpha=1,\dots,N_C$ enumerates the concepts.

In principle, any simply connected region could be used to define $C_{\alpha,k}$, but for simplicity, hyper-ellipses are employed in each description space. 

It is important to note that not all samples in the set of candidates $C_{\alpha, k}$ will eventually belong to the concept $\alpha$ but only a fraction of those samples make up the concept. 
A concept $C_\alpha$ is defined as a set of samples which comprises all those samples $x$ that are associated with one and only one concept $\alpha$ in all description spaces $k$ simultaneously and thus do not belong to any other concept. 
Formally, the set of data samples belonging to a concept $\alpha$ can be expressed as
\begin{equation}\label{eq:concept_c_alpha}
    C_\alpha = \Big\{  x \in \bigcap_{k=1}^{N_{DS}} C_{\alpha, k} \Big| x \not \in C_{\beta, k} ; \forall k,\beta\neq\alpha\Big\},
\end{equation}
where $k,l=1,\dots,N_{DS}$ denote description spaces and $\alpha,\beta=1,\dots, N_C$ denote concepts. 
As noted earlier, an important distinction has to be made between the samples in the set $C_{\alpha, k}$ and in the concept  $C_\alpha$. 
The former comprises all samples in description space $k$ which lie inside the hyper-ellipse for concept $\alpha$. 
This includes all those samples which also belong to other concepts $C_{\beta, k}$ in the same description space or which are not included in $C_{\alpha, l}$ of other description spaces.  
In stark contrast, the samples defining the concept $C_\alpha$ only comprise those samples exclusively associated to $C_{\alpha, k}$ as well as $C_{\alpha, l}$ for all other spaces.

As mentioned earlier, the ability to integrate user preferences into the definition of concepts is highly desirable.
Despite being rather hard to formulate in general, users can typically give exemplary data samples which are roughly in line with their preference \cite{wangPreference2017,raySOI2022}.
This fact is utilized and user preferences is presented as a set of data samples which are of interest to the engineer. 
It is therefore assumed that the user selects few samples of interest in each description space.
The method then aims at incorporating these samples of interest into the concepts. 
The set of preference samples is denoted as $P$ which contains $N_{Pref}$ data samples.

\begin{figure*}[!ht]
\centering
\includegraphics[width=.95\textwidth]{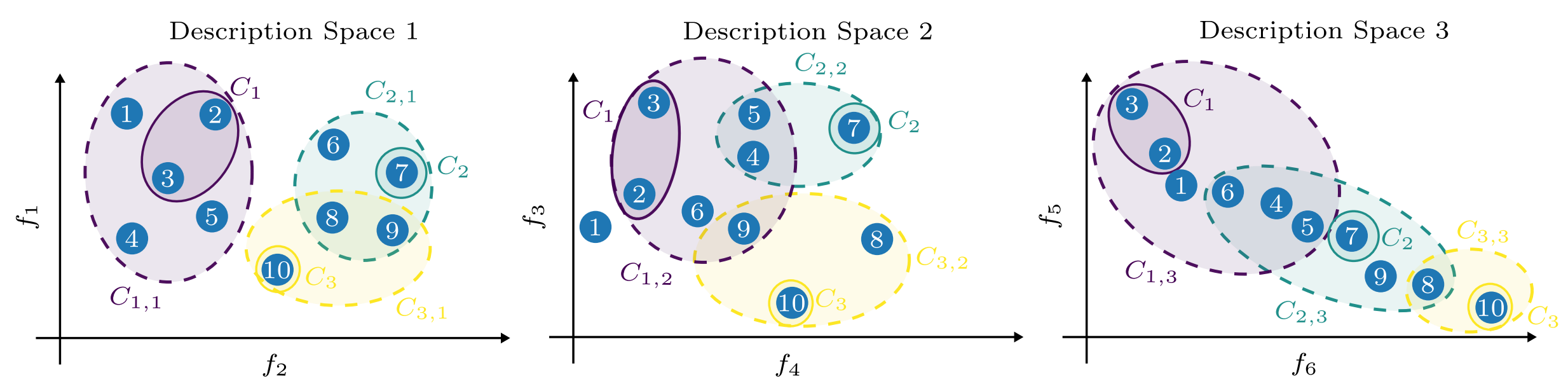}
\caption{Example dataset and distribution of concepts. A dataset containing six descriptive features ($f_1$ to $f_6$) is split into three description spaces. 
Data samples inside  the dashed light-colored ellipses are candidates for the corresponding concept in each description space, while the stronger colored ellipses indicated the actual concepts.} \label{fig:dataset_example}
\end{figure*}

An example dataset to demonstrate the definitions and the concept identification process is illustrated in Figure~\ref{fig:dataset_example}.
The dataset contains ten samples where each is described by six features $f_1$ to $f_6$. 
The features are split into three description spaces with two features each.
The figure also contains proposed concept regions as colored ellipses, where the points inside each ellipse are considered candidates for the corresponding concept.
In the shown example, the samples $x_1$ to $x_5$ lie inside the purple ellipse in description space one and are thus candidates for the corresponding purple concept one, 
i.e., $C_{1,1} = \left\{ x_1,x_2,x_3,x_4,x_5 \right\}$.
Similarly, the candidate sets for the purple concept one in the other spaces are $C_{1,2} = \left\{ x_2,x_3,x_4,x_5,x_6,x_9 \right\}$ and $C_{1,3} = \left\{ x_1,x_2,x_3,x_4,x_5,x_6 \right\}$.
From these candidates the actual samples belonging to concept one can be determined: only samples $x_2$, $x_3$, $x_4$ and $x_5$ are candidates for concept 1 in all description spaces, but samples $x_4$ and $x_5$ are also associated with other concepts in description spaces two and three, and should therefore be excluded as well.
As a result, the purple concept contains the samples $C_{1} = \left\{ x_2,x_3 \right\}$ only. 
Similarly, the other concepts are $C_{2} = \left\{ x_7 \right\}$, and $C_{3} = \left\{ x_{10}\right\}$. 
In this example it can be observed, that the concept distribution could be improved, for example just by changing the ellipses of the green concept such that they does not include samples $x_8$ and $x_9$ in description space one and three.   
The concept identification problem amounts to finding the optimal ellipses in each description space which partition the dataset into meaningful and consistent concepts which cover as many data samples as possible.

In order to assess the quality of a given definition of a set of concepts (for example, specified by their ellipses in each description space), a novel objective numerical concept quality measure (CQM) $Q$ is defined.
The quality of a set of multiple concepts is calculated  as the product of individual metrics $Q_\alpha$ for each individual concept $\alpha$ as
\begin{equation}\label{eq:q}
 Q=\prod_\alpha^{N_C}Q_\alpha\:.
\end{equation}
The individual metric $Q_\alpha$ for each concept is defined as 
\begin{equation}\label{eq:q_alpha}
 \begin{aligned}
  Q_\alpha = & \left(\prod_{k}^{N_{DS}} \sqrt[N_{DS}]{\frac{| C_\alpha |} {|C_{\alpha,k} |}} \right)\cdot F \left( \frac{|C_{\alpha} |} {N_\mathcal{D}}, s \right) \\
  & \cdot F \left( \frac{|P_{\alpha}|} {|P| }, p \right)\:,
  \end{aligned}
\end{equation}
where $|C|$ denotes the size of set $C$, i.e. the number elements in $C$. 
Despite the seemingly complex appearance of Eq.\eqref{eq:q_alpha}, the idea behind the CQM is quite simple.  
The measure quantifies how many samples actually belong to a concept $C_\alpha$, relative to the number of samples $C_{\alpha, k}$ inside the hyper-ellipses in each description space $k$. 
This is accounted for by the first factor involving the fraction, and the geometric mean of the fractions is taken over the contributions of each description space. 
Since $|C_\alpha|\leq| C_{\alpha, k}|$, the upper bound for this factor is one for an ideal concept and amounts to zero in the worst case. 
Generally, large overlap between the concept regions leads to low values for the fraction and vice versa. 

The measure also includes two additional factors which involve the helper function 
\begin{equation}
\label{eq:scaling_function}
  F(x,y) = \begin{cases}
			\sqrt{1 - \left( \frac{x - y}{y} \right) ^2}, & \text{if $x < y$,}\\
			\quad 1, & \text{if $ y < x < 1-y$}, \\
			\sqrt{1 - \left( \frac{x -1 + y}{y} \right) ^2}, & \text{if $x > 1-y$}.
		\end{cases}
\end{equation}
The first factor $ F ( |C_{\alpha, k} |/N_\mathcal{D}, s ) $, where $N_\mathcal{D}$ is the total number of samples in the dataset, allows for the adjustment of the desired minimal and maximal size of each concept. 
It reduces the quality measure as soon as the size of the concept is not in the range $s\leq |C_{\alpha} |/N_\mathcal{D} \leq1-s$, with a freely choseable parameter $0\leq s\leq \tfrac{1}{2}$.

The second factor reduces the quality measure if a concept includes too few preference samples.  
The factor is unity, if the number preference samples associated with concept $\alpha$, denoted by $P_{\alpha}$ relative to the total number of preference samples, denoted by $P$, is in the range 
$p\leq |P_{\alpha}| /  |P| \leq 1-p$,  with  a freely choseable parameter $0\leq p\leq\tfrac{1}{2}$. 
Otherwise the factor is reduced and thereby the quality of the concepts is also reduced. 
With this it is enforced that each concept includes not too few but also not too many preference samples.

In order to elucidate the CQM, it is calculated for the example dataset of Figure~\ref{fig:dataset_example}. 
For brevity, the preference samples are omitted and $s=0.15$ is used in the size penalty function.  
The score $Q_{1}$ for concept $1$ evaluates to
\begin{equation}
 \begin{aligned}
      Q_1 &  = 
     \left(\prod_{k}^{3} \sqrt[3]{\tfrac{| C_1 |} {|C_{1 k} |}} \right)  \cdot F \left( \tfrac{|C_{1} |} {N_\mathcal{D}}, s \right)  \\
     & =\sqrt[3]{\tfrac{| C_1 |} {|C_{1,1} |}} \cdot \sqrt[3]{\tfrac{| C_1 |} {|C_{1,2} |}} \cdot \sqrt[3]{\tfrac{| C_1 |} {|C_{1,3} |}}  \cdot F \left( \tfrac{2} {10}, 0.15 \right) \\
      & = \sqrt[3]{\frac{2} {5}\cdot\frac{2} {6}\cdot\frac{2} {6}}  \cdot 1  = 0.354\:.
 \end{aligned}
\end{equation}
Analogously, the metric scores for the other concepts are 
\begin{equation}
 \begin{aligned}
      Q_2 & =    \sqrt[3]{\tfrac{1} {4}\tfrac{1} {3}\tfrac{1} {6}} \cdot F \left( \tfrac{1} {10}, 0.15 \right) = 0.240 \cdot0.94= 0.226,\\
      Q_3 & =    \sqrt[3]{\tfrac{1} {2}\tfrac{1} {3}\tfrac{1} {2}} \cdot F \left( \tfrac{1} {10}, 0.15 \right) = 0.437 \cdot0.94= 0.411 \:.
 \end{aligned}
\end{equation}
The metrics $Q_2$ and $Q_3$ are penalized since both concepts contain less than 15\% of all samples in the dataset.
The  overall CQM evaluates to 
\begin{equation*}
 \begin{aligned}
      Q & = Q_1\cdot Q_2\cdot Q_3= 0.033\:
 \end{aligned}
\end{equation*}
which is rather poor. 
This is in line with the expectation for this example, as only four out of ten data samples can be consistently attributed to concepts and the overlap between the concept candidates is large.

In summary, the metric provides a numerical measure between zero and one which quantifies how good the association of samples with multiple concepts in multiple descriptions spaces is. 
The optimal value of $Q=1$ can only be attained if all data samples can be unambiguously assigned to a concept. 
The numerical value of the measure is reduced if the size of the concepts is not within a reasonable, user-defined range. 
Similarly, the measure is also reduced if the concepts do not include enough samples that are of particular interest to the user.
These features are explicitly valuable when the metric is used to optimize the distribution of concepts in a dataset.
For example, when a description space consists of competing design objectives, high performing (non-dominated) solutions can be considered as preferred samples.
This would drive the concept optimization towards high performing solutions.
In contrast to~\cite{Graening2014}, where the hypervolume of the concept in an objective space is considered to build the utility metric, the proposed measure allows a higher diversity when choosing samples of interest and multiple directions to drive the identification process.

Such a numeric measure is very useful when multiple alternative partitions of one dataset into concepts are proposed. 
The numerical measure provides a direct quantification of concept quality and can thus be used to select the best definition of the concepts. 
Therefore, it can also be used to find optimal concepts by formulating the concept identification task as a numerical optimization problem where the CQM is used as an objective function that is to be maximized. 
The regions which define the candidates for each concept in each description space are then parametrized and the optimization approach targets to find the optimal parameters, thereby defining the optimal shapes of the concept regions. 
In all experiments in this work, hyper-ellipses are used as shapes for the concept candidate regions and their parameters are optimized to maximize the CQM.

\section{Concept Identification Experiments}\label{sec:experiments}

To demonstrate the necessity for the proposed metric, an example case is discussed in detail.
The example illustrates the difference between considering multiple description spaces in parallel, in comparison to considering only one.
To show the applicability of the methods towards realistic use-cases, two experiments with a real-world-inspired airfoil dataset are further conducted and analyzed.

\subsection{Demonstration Example}\label{subsec:demonstration_example}

\begin{figure*}[!ht]
    \centering
    \begin{subfigure}{.3\textwidth}
      \centering
      \includegraphics[width=0.96\linewidth]{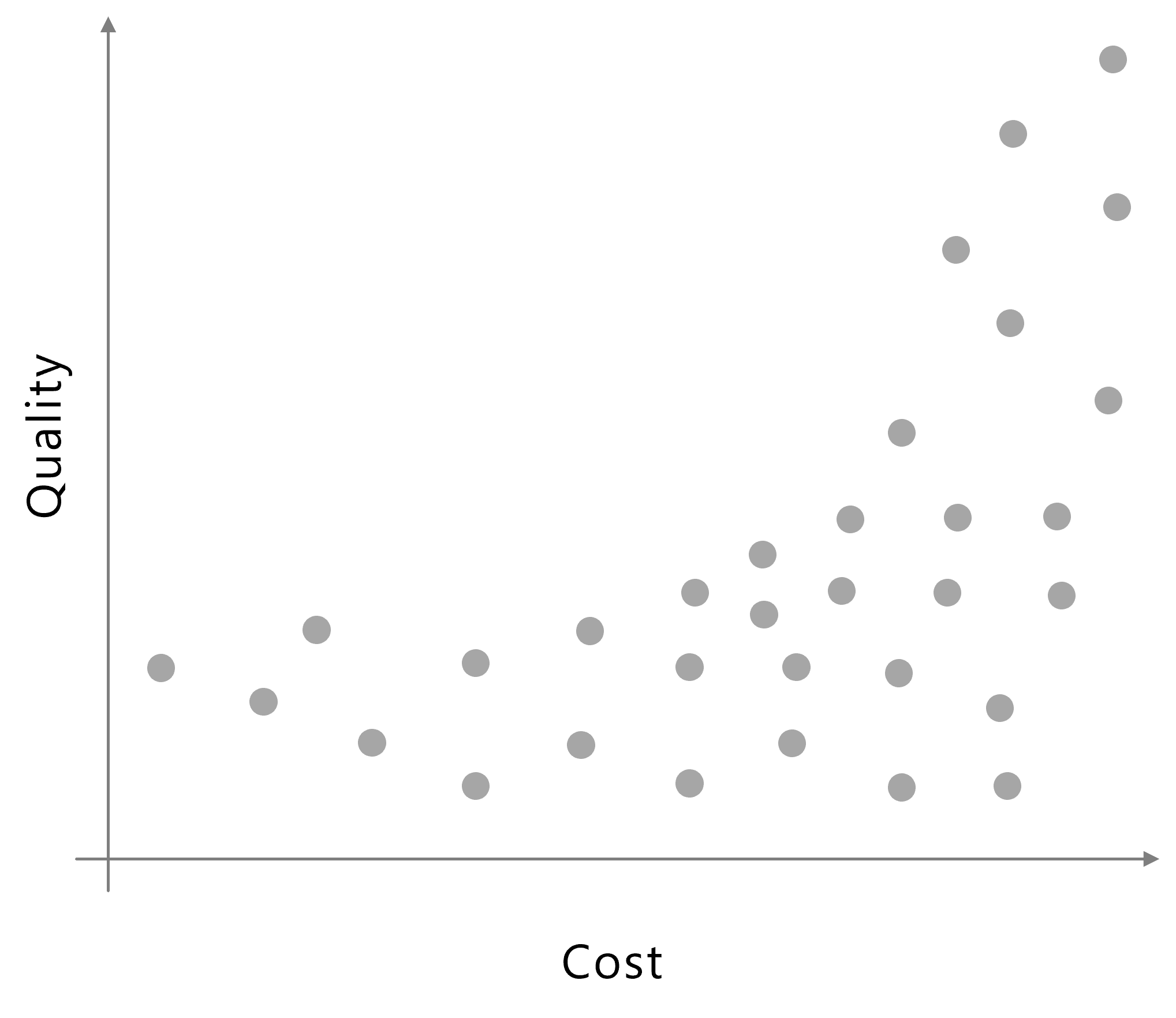}
      \caption{Dataset}
      \label{fig:e1}
    \end{subfigure}
    \begin{subfigure}{.3\textwidth}
      \centering
      \includegraphics[width=0.96\linewidth]{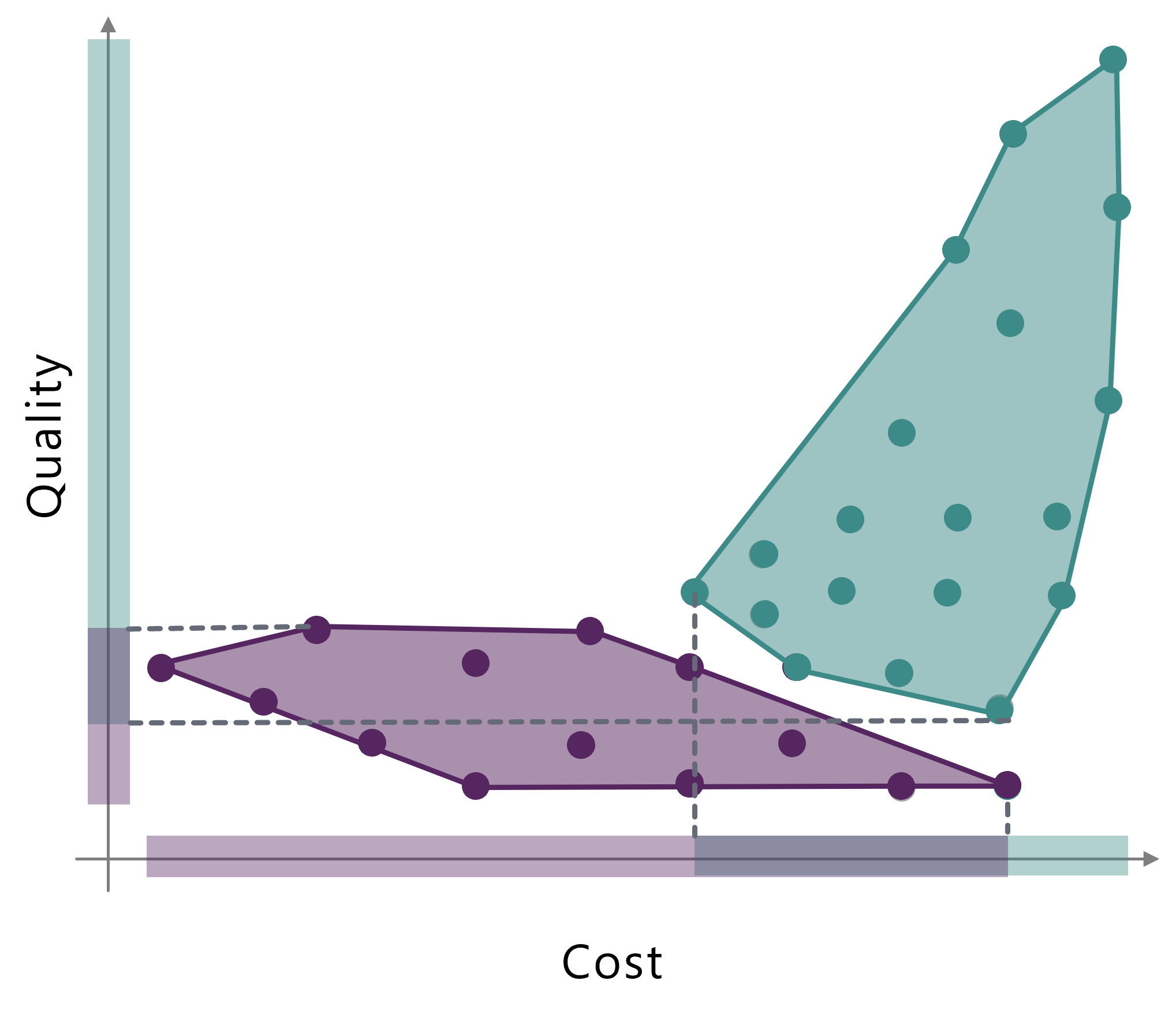}
      \caption{Clustering based concepts}
      \label{fig:e2}
    \end{subfigure}
    \begin{subfigure}{.3\textwidth}
      \centering
      \includegraphics[width=.96\linewidth]{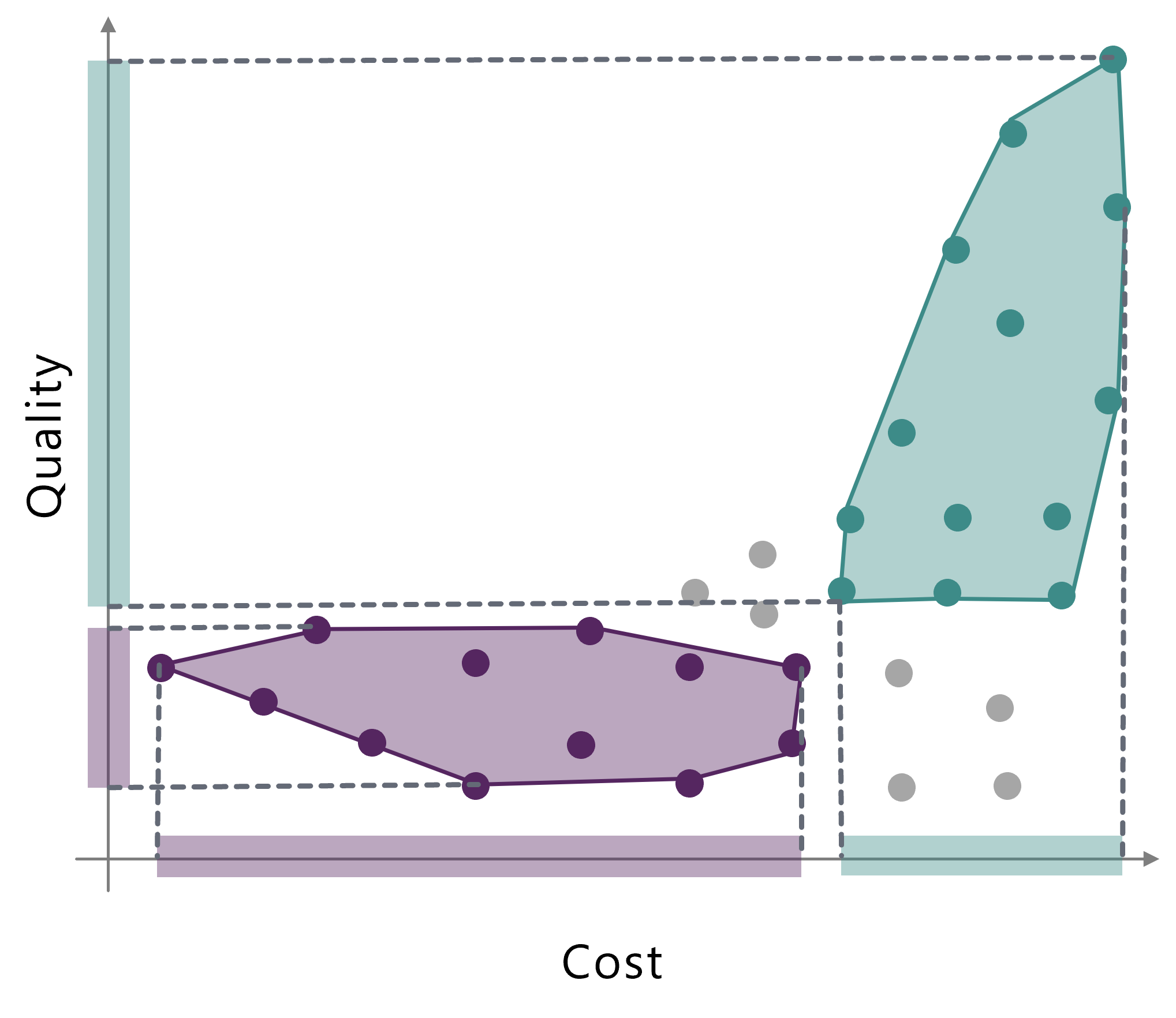}
      \caption{Concept identification}
      \label{fig:e3}
    \end{subfigure}
\caption{Multi-space concept identification demonstration example. (a) The dataset is defined by two features, cost and quality. (b) Two concepts (purple and green) are determined by some clustering of the two-dimensional feature space. The colored bars on the axis indicate the extension of the concept on each feature axis. 
The overlap is indicated in grey. (c) Two concepts are identified based on two separate one-dimensional description spaces. The dotted shaded areas depict the projections of the concepts region to  the feature axis.} \label{fig:demonstration_experiment}
\end{figure*}

Consider a dataset containing design samples where each sample has an associated cost (e.g., to manufacture) and quality. 
Each of these features is described by one scalar value and therefore the dataset can be depicted in a two-dimensional feature space (Figure~\ref{fig:demonstration_experiment}(a)).
If the samples should be divided into two concepts, the result could be obtained by a regular clustering approach (Figure~\ref{fig:demonstration_experiment}(b)).
In this case, a meaningful interpretation of the concepts is rather difficult.
The samples from the green concept are of high quality and have large cost, while the samples associated with the purple concept have much lower cost but also lower quality. 
Therefore, the green and purple concepts would correspond to a \enquote{high-end} and a \enquote{low-end} concept, respectively.
However, both concepts also include samples which do not align with such an interpretation: the green concept includes samples with higher cost but lower quality than  samples in the purple concept, and are therefore likely uninteresting for a user, as they would not be considered \enquote{high-end} samples.
A proper \enquote{high-end} concept would suggest that all samples should have at least a minimum level of (high) quality compared to the other concept(s).

On the other hand, the outcome is more intuitive if the concept identification process is conducted using two separate description spaces. 
The quality serves as one one-dimensional description space and cost as another one-dimensional description space (Figure~\ref{fig:demonstration_experiment}(c)).
In this case, the process identifies concepts that are separable with respect to both features.
If the concepts are projected in the two-dimensional space of cost and quality, the concept identification can be understood as a division of the samples based on orthogonal lines.
All samples within the green concept show a higher cost value than the purple samples, but also a consistently higher quality value.

\subsection{Airfoil Experiments}\label{subsec:ex_airfoil}

\begin{figure*}[!ht]
\centering
\includegraphics[width=.96\textwidth]{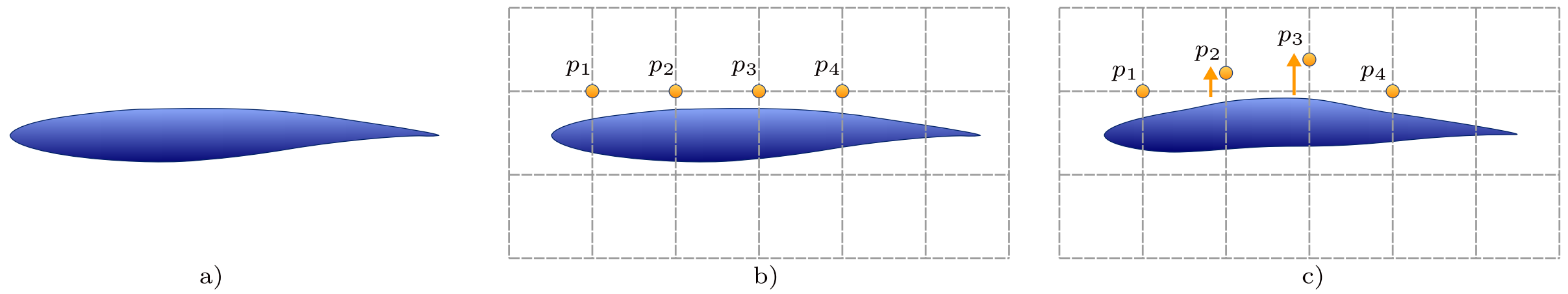}
\caption{Creation of the Airfoil dataset. The two-dimensional airfoil profiles are created using free-form deformation. The RAE2822 base profile of panel (a) is embedded in a lattice (b). Four of the lattice control points are defined as control parameters and varied to create deformed profiles (c). The dataset contains 2503 deformed airfoil profiles.} \label{fig:airfoil_deformation}
\end{figure*}

A dataset derived from the aerodynamic evaluation of two-dimensional airfoil profiles is used in the following.
The profiles are constructed by applying free-form deformations (FFD)~\cite{Sederberg1986FreeformDO,Sieger2012} to a baseline RAE2822 profile~\cite{rae}, see Figure~\ref{fig:airfoil_deformation} for the deformation setup.
For each generated airfoil shape, up to 15 features are recorded and the total number of data samples is $N_\mathcal{D}=2503$. 
The design parameters which define the shape are given by the four FFD deformation parameters, each specifying a deformation at a predefined location. 
Additionally, the position of the camber line is calculated for five different locations along the airfoil, thereby providing five geometric features of the airfoil.
The aerodynamic behavior of each profile is evaluated by calculating the drag and lift coefficients of the airfoil (see, e.g.~\cite{andersonAero}) using the computational fluid dynamics (CFD) solver OpenFOAM\footnote{www.openfoam.org}. 
Please note that the reported lift and drag coefficients $C_{lift}$ and $C_{drag}$ are normalized to the dataset, that is the calculated coefficients are always transformed to be in the interval $[0,1]$. 
The aerodynamic performance is evaluated for fixed airspeed but at three different boundary conditions specified by three different angles of attack $\alpha=0^\circ$, $\alpha=1^\circ$ and $\alpha=3^\circ$.
Three different angles of attack are considered to reflect characteristic operating conditions such as cruise flight, landing or take-off, and typically a tradeoff between objectives is observed. 
The dataset was created by performing multiple evolutionary optimization runs which targeted the minimization of the drag of the airfoils while meeting requirements on the lift values at various conditions.

The concept identification is carried out as an optimization where the CQM of Eq.\eqref{eq:q} is used as the objective function to be maximized.
For this, the covariance matrix adaptation evolutionary strategy (CMA-ES)\cite{hansenCMA,Hansen2001} is employed to find an optimal distribution of concepts by maximizing the CQM.
Each concept is parametrized by a hyper-ellipsoid in each description space, where each data sample inside the ellipsoid is a candidate for the corresponding concept.
For an $n$-dimensional description space, a hyper-ellipsoid has $n(n+3)/2$ parameters which need to be optimized by the search.

\begin{table*}[!ht]
\centering
\caption{Details of the experimental setup and optimization configuration}
\begin{tabularx}{0.96\textwidth}{lYYY}  
\toprule
Experiment & Airfoil two spaces & Airfoil four spaces & Airfoil five spaces \\
\midrule
\# description spaces  $N_{DS}$ & 2 & 4 & 5\\
\# features per space & 4, 6 & 4, 2, 2, 2 & 4, 5, 2, 2, 2 \\
\# concepts to be identified $N_{C}$  & 3 & 3 & 3 \\
optimization algorithm & CMA-ES & CMA-ES & CMA-ES\\
\# search parameters  & 123 & 87 & 147 \\
\# individuals & 200 & 200 & 200\\
\# generations & 320 & 320 & 320\\
size scaling parameter $s$ & 0.01  & 0.01 & 0.01\\
preference scaling parameter $p$  & - & 0.01 & 0.01\\
\bottomrule
\end{tabularx}
\label{tab:experiment_setup}
\end{table*}

\subsubsection{Airfoil in two Description Spaces}

\begin{figure*}[!ht]
\centering
\includegraphics[width=.96\textwidth]{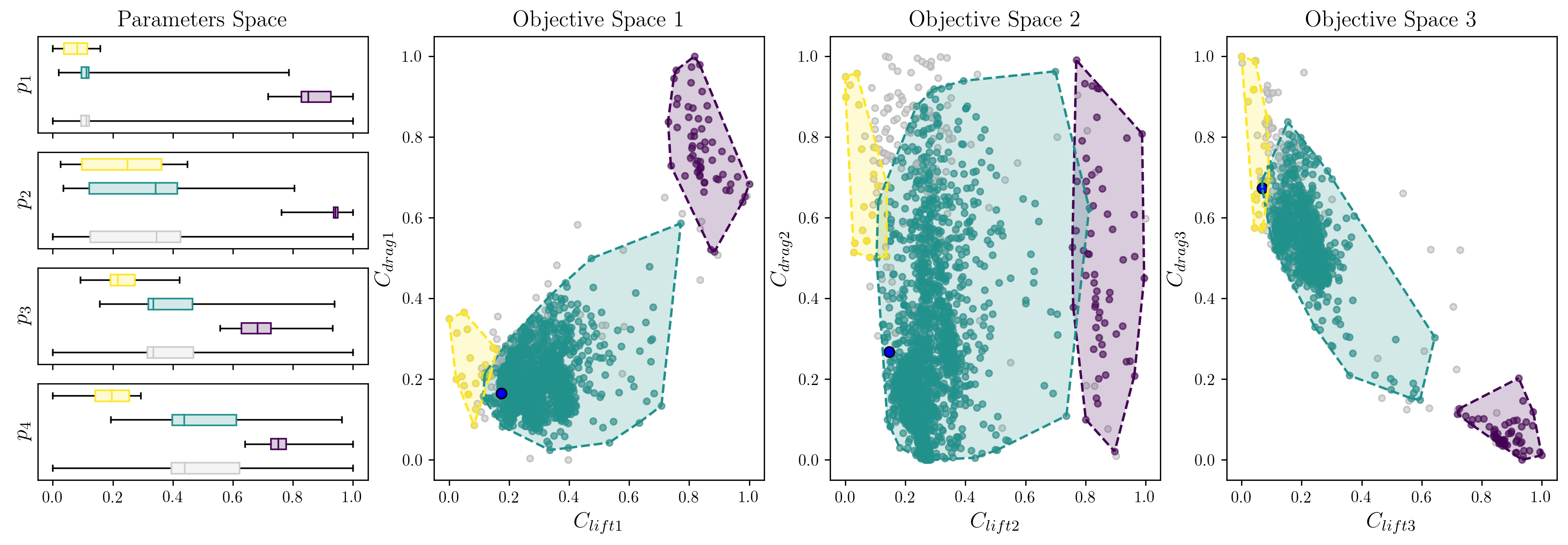}
\caption{Concept identification result for the airfoil dataset in two description spaces. The four-dimensional design parameter space is shown as projections  into the parameters, while the six-dimensional objective space is shown as projections into three separate two-dimensional spaces, one for each operating condition, i.e.\ angle of attack. The concepts (purple, green, yellow) are depicted as the convex hull of their respective associated samples in the 2-dimensional spaces. A significant amount of overlap between the concepts is visible. E.g., the blue circle marks a sample that can be associated with the yellow and green concept in objective space 3.} \label{fig:AF_2spaces_all_OS}
\end{figure*}

In the first experiment, the four deformation parameters and the six objective values are selected and concepts are determined based on these $N_F=10$ features. 
The partitioning into design parameters and objective values is typical for engineering approaches as they naturally represent semantically separated feature types. 
In the formulation, this corresponds to $N_{DS}=2$ description spaces with four and six dimensions, representing the design parameters and the objectives (drag and lift for three angels of attack), respectively. 

The experiment aims at identifying $N_C=3$ distinct concepts, which are represented as a set of $3 \cdot (4 \cdot (4+3)/2 + 6 \cdot (6+3)/2)=123$ parameters for the hyper-ellipses to be determined by the optimization.
In this example, no preference samples are included.
The experimental setup and optimization configuration is given in Table~\ref{tab:experiment_setup}.

Figure~\ref{fig:AF_2spaces_all_OS} depicts the concepts obtained with a typical optimization run, where the three identified concepts are marked in yellow, purple, and green. 
The grey points are data samples which are not associated with any concept. 
The parameter space is given as a box-plot for each deformation parameter, while for the six dimensional objective space the projections onto three two-dimensional spaces are shown, each one spanned by the performance values for one angle of attack. 
Note that the dataset is normalized such that each parameter and objective value lies in the range between zero and one. 

The results look quite intuitive as the three concepts correspond roughly to low, medium, and high lift airfoils at all operating conditions, i.e., angles of attack. 
Inspecting the relation between objective space and parameter space also reveals that the purple high lift concept has high values for the deformation parameters which translates to high camber airfoils. 
Therefore the concepts are roughly identifying meaningful groups of similar designs.   
However, even though there is no overlap in the concepts in each description space, this is not true for the  two-dimensional projections onto the shown subspaces. 
This is especially unfavorable for the objective spaces.
Here, the green and the yellow concept overlap. 
Similarly, the purple concept is overlapping with the green concept in objective space 2. 
This undermines a consistent interpretation of the concepts in terms of high, medium, and low lift airfoils, similar to the example of Section~\ref{subsec:demonstration_example}. 
Therefore, selecting samples from each concept does not guarantee that these samples can also be meaningfully categorized when considering the objective spaces separately. 

One could argue, that, in this example the overlap of the concepts, and with it the probability of selecting samples from these overlapping regions, seems rather small, and that therefore this effect could be ignored by just discarding the overlapping samples.
But this is not the case in general. 
For other examples, the overlap could be much larger and depending on the distribution of data samples, the effect would be much more pronounced. 
Also, in objective spaces with a dimension higher than three, understanding the concept overlap by simply visualizing the concepts is very hard. 
Therefore, an objective numerical measure as proposed in this work, to achieve non-overlapping concepts in multiple description spaces, is highly desired.

\subsubsection{Airfoil in Multiple Description Spaces}\label{subsubsec:ex_airfoil_4spaces}

\begin{figure*}[!ht]
\centering
\includegraphics[width=0.96\textwidth]{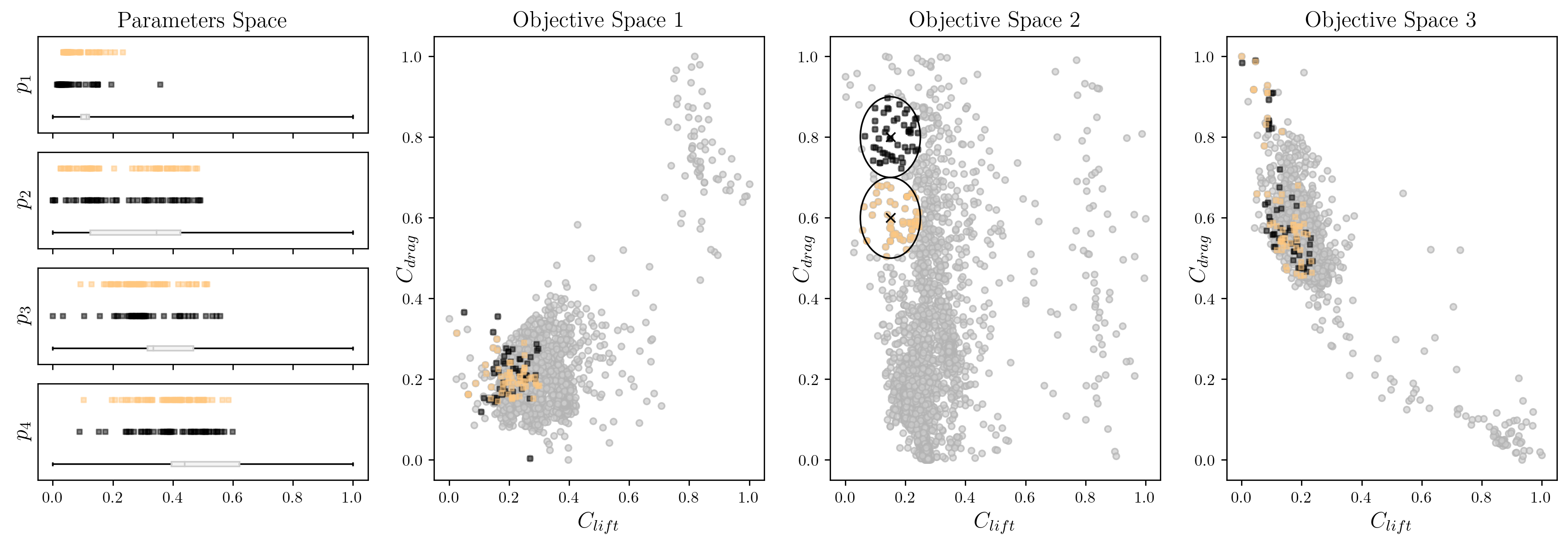}
\caption{Structure of the dataset. Data samples in two regions of objective space 2 are  highlighted in black and orange brown, and shown in all other description spaces. While the sets are similar in objective space 2 and do not overlap with each other, they show a significant amount of overlap in the other spaces and are generally more spread out.} \label{fig:AF_Circles}
\end{figure*}

To illustrate the advantages of the approach described in section~\ref{sec:methods}, a second experiment with the same airfoil dataset was conducted.
In this case, the ten features are divided among four description spaces.
The four parameters form the parameters space, as in the previous scenario.
In contrast to the previous experiment, the combinations of drag and lift coefficient for the respective angle of attack each form one additional description space.
The scenario is hence constructed out of four description spaces, one four-dimensional space and three two-dimensional spaces.

Before identifying concepts based on the proposed CQM, the structure of the dataset is investigated further. 
Figure~\ref{fig:AF_Circles} shows the dataset projected into the four description spaces and two groups of data samples are colored in black and orange. 
The groups are composed of all samples inside two adjacent ellipses in objective space 2. 
As can be seen, these groups which are disjunct in objective space 2  have strong overlap in the other  description spaces. 
In the present example, such a behaviour results from the complex fluid flow field around the airfoils obtained by the CFD solver. 
Similar drag and lift at one angle of attack can be obtained with different geometries, depending on the mechanism and the geometric features of the airfoil which influence these values. 
For example, the trailing edge of the airfoil, which is mostly influenced by $p_4$, has a strong effect on the lift, but similarly, the mean camber (i.e. the curvature of the airfoil), which is determined by a more coordinated movement of all parameters,  has an influence. 
At a different angle of attack, the influence of the specific camber and trailing edge shape might be quite different, therefore leading to differing values of drag and lift at other angles of attack. 
Therefore, the observed structure of the dataset shown in Figure~\ref{fig:AF_Circles} is understandable and can be expected to be the generic behavior for engineering applications. 
This is especially true, in situations where only incomplete feature sets are included in the dataset, because discriminating aspects are not reflected in the data.

\begin{figure*}[!ht]
\centering
(a)\phantom{FFFFFFFFFFFFFFFFFFFFFFFFFFFFFFFFFFFFFFFFFFFFFFFFFFFFFFFFFFFFFFFFFFFFF}\\
\includegraphics[width=.96\textwidth]{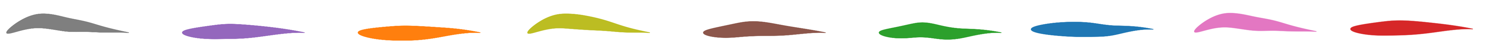}\\
(b)\phantom{FFFFFFFFFFFFFFFFFFFFFFFFFFFFFFFFFFFFFFFFFFFFFFFFFFFFFFFFFFFFFFFFFFFFF}\\[-0mm]
\includegraphics[width=.96\textwidth]{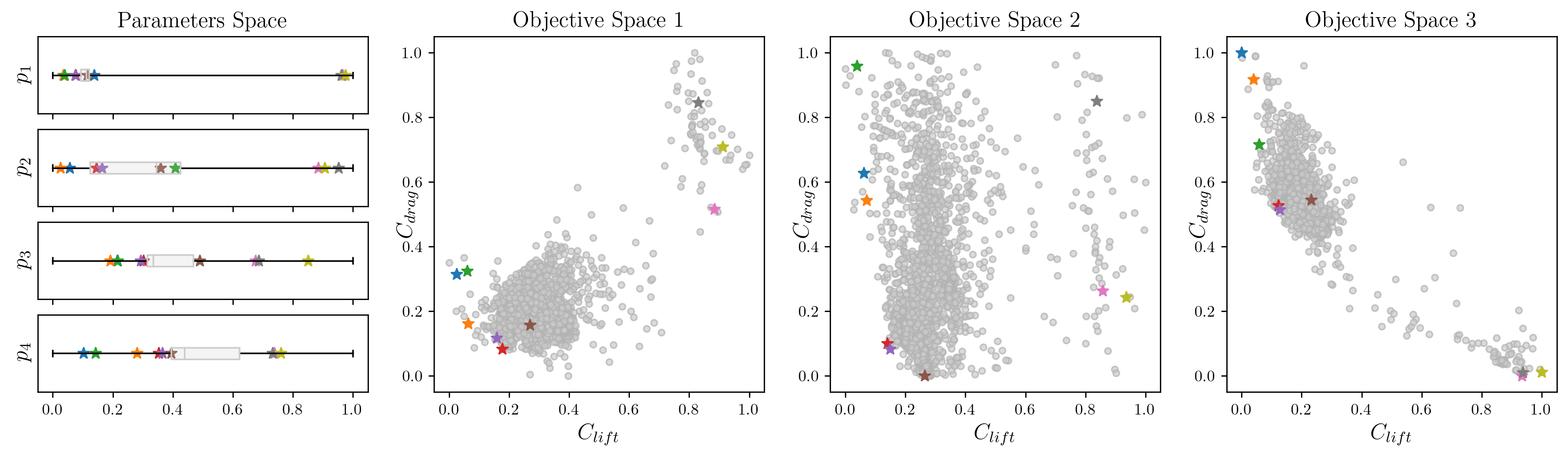}
\caption{Solutions of interest. (a) Nine specific airfoils which serve as solutions of interest. (b) The solutions of interest are indicated as colored stars in the airfoil  dataset in four description. The colors of the point markers match the colors in panels (a). } \label{fig:AF_specific_samples}
\end{figure*}

As discussed earlier, incorporating user-preference into the concept definitions is a highly desirable aspect. 
This is achieved by first letting the user choose a small set of designs that are of particular interest.
These airfoil profiles as well as their location in the dataset are shown in Figure~\ref{fig:AF_specific_samples}.
In this example, the design samples of interest are a mix of designs that lead to high lift under take-off conditions, low drag for small angles of attack, and good tradeoff solutions. 
The idea is that the identified concepts incorporate these samples into concepts, without the need for the use to specify any association with concepts from the start.

The identification of reasonable concepts is again carried out as an optimization process, where hyper-ellipses represent the concepts in each description space and the optimization maximizes the CQM.
The experiment aims at identifying $N_C=3$ distinct concepts in the given four description spaces which results in $87$ search  parameters.
The experimental configuration is again summarized in Table~\ref{tab:experiment_setup}.

\begin{figure*}[!ht]
\centering
\includegraphics[width=.96\textwidth]{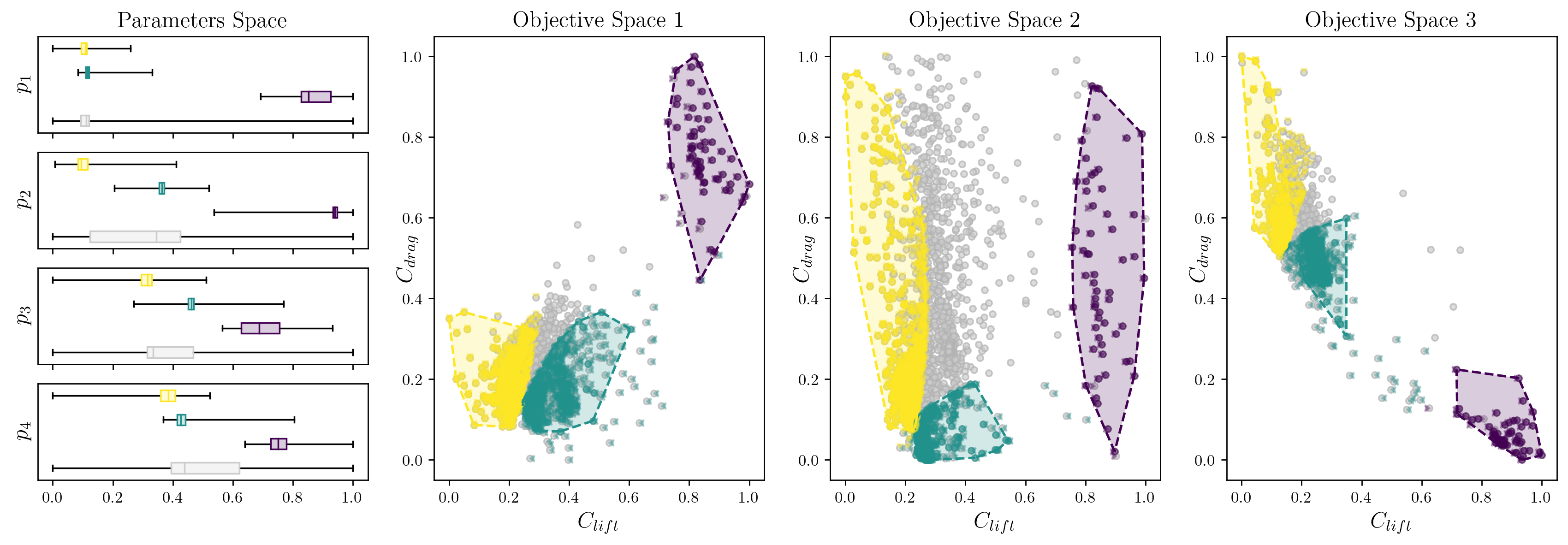}
\caption{Result of the concept identification process in four description spaces. Three concepts (purple, green and yellow samples) are identified that each cover parts of each description space. 
The colored regions indicate the concept candidate regions $C_{\alpha,k}$. Inside each concept candidate region the samples actually belonging to concept $C_\alpha$ are also colored, while  there are still data samples in grey which are not associated with the corresponding concept.  } \label{fig:AF_Result_4Spaces}
\end{figure*}

\begin{figure}[!ht]
\centering
\includegraphics[width=0.3\textwidth]{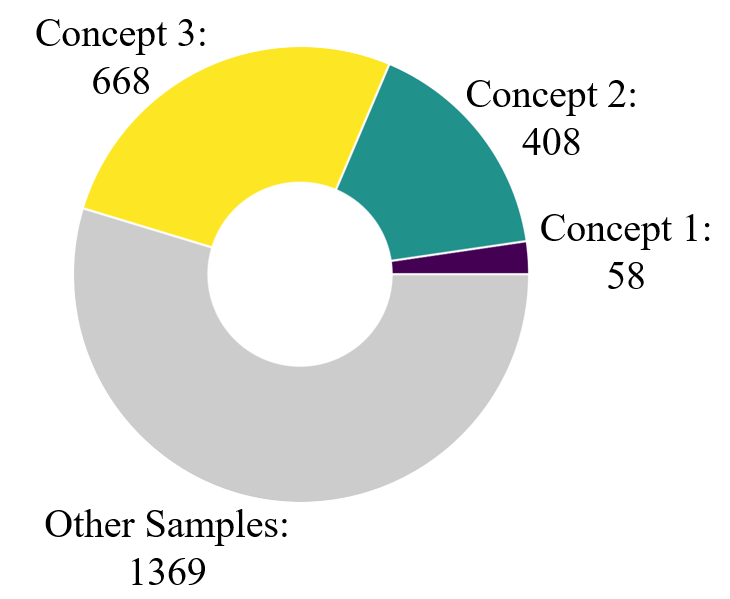}
\caption{Concept sizes. Number of samples for each concept for the airfoil experiment using four description spaces.} \label{fig:AF_Quality_4Spaces}
\end{figure}

The results of a concept identification optimization is shown in Figure~\ref{fig:AF_Result_4Spaces}. 
The identified concepts seem to be quite similar to the concepts found when considering only two description spaces shown in Figure~\ref{fig:AF_2spaces_all_OS}. 
However, the green concept is much smaller and concentrates more on the low drag region in objective space 2. 
It also covers a much narrower region in the design parameter space, which is desirable from the engineering point of view.  
However, the concepts cover large regions of each description space, which reflects the design goal to associate as many data samples with a concept as possible. 
Still, many data samples are not associated with any concept, which can be guessed from Figures~\ref{fig:AF_Result_4Spaces} and which is obvious when inspecting the numbers of samples in each concept (Figure~\ref{fig:AF_Quality_4Spaces}). 
In the ladder figure, it can be also observed that the number of samples associated with each concept varies considerably, but the smallest concept (purple) still contains enough data samples to be considered meaningful.

\begin{table}[!ht]
\centering
\caption{Metric scores for the identified concepts in four description spaces. 
The total CQM for these concepts is $Q=0.54$.}
\begin{tabularx}{0.96\linewidth}{lcc}
\toprule
& \# Samples & CQM per concept $Q_\alpha$ \\
\midrule
\rowcolor{lightPurple}
Concept 1 & 58 & 0.95 \\
\rowcolor{lightGreen}
Concept 2 & 408 & 0.68\\
\rowcolor{lightYellow}
Concept 3 & 668 & 0.84 \\
\bottomrule
\end{tabularx}
\label{tab:metric_values_4spaces}
\end{table}

The values of the CQM per concept $Q_\alpha$ of Eq.~\eqref{eq:q_alpha} are reported in Table~\ref{tab:metric_values_4spaces}. 
The scaling factors for concept size and preference samples (involving the scaling function of Eq.~\ref{eq:scaling_function}) were always 1 (not shown). 
While the CQM score for the purple concept 1 is close to one, which indicates that most of the data samples inside the ellipses of that concept are also actually associated with that concept, the other scores are smaller. 
In particular, the score for the green concept 2 is only 0.68 which indicates that many samples inside the green ellipses are not associated with that concept. 
This is in line with the previous observation, that the proximity of samples in one description space not necessarily implies the proximity in other description spaces due to the complexity of the underlying engineering problem or an incomplete feature set. 
However, the proposed CQM and concept identification process are still able to find extended concepts extending over large regions of the dataset where the samples associated with each concept have the desired property of feature similarity in each description space.
Also, the identified concepts successfully incorporated the predetermined solutions of interest.

\subsubsection{Airfoil in Multiple Description Spaces Including Features}

\begin{figure}[!ht]
\centering
\includegraphics[width=0.45\textwidth]{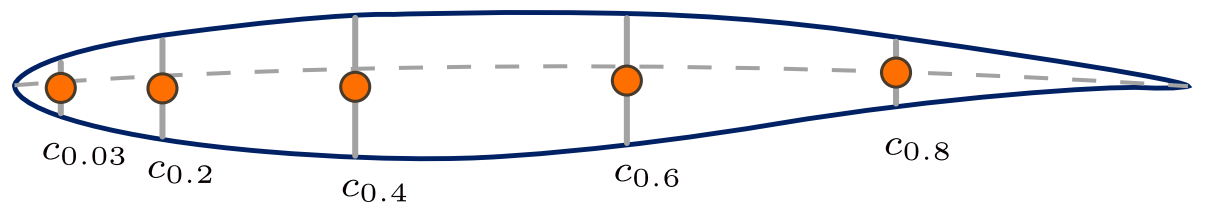}
\caption{Additional features for the airfoil dataset. The height of the camber line at five different position is calculated from each profile and defines an additional description space for the dataset.} \label{fig:airfoil_features}
\end{figure}

In many contexts, additional geometric features of the designs are considered.  
In order to demonstrate how the proposed method can incorporate such additional features, the current airfoil dataset is extended with additional geometric features derived from the airfoil profiles. 
For each profile, the $y$-position of the camber line (i.e. the center line of the airfoil) is calculated at five predefined positions along the airfoil (Figure~\ref{fig:airfoil_features}).

These geometric features are used as an additional description space and are incorporated in the concept identification process.
Therefore, the following five description spaces are considered: the parameter space, the geometric feature space and the three objective spaces.
The same samples of particular interest as in section~\ref{subsubsec:ex_airfoil_4spaces} are included as user-preferences.

The optimization process for concept identification uses again a hyper-ellipse representation with now $147$ search parameters for identifying $N_C=3$ distinct concepts. 
The experimental configuration is again according to Table~\ref{tab:experiment_setup}.

\begin{figure*}[!ht]
\centering
\includegraphics[width=.96\textwidth]{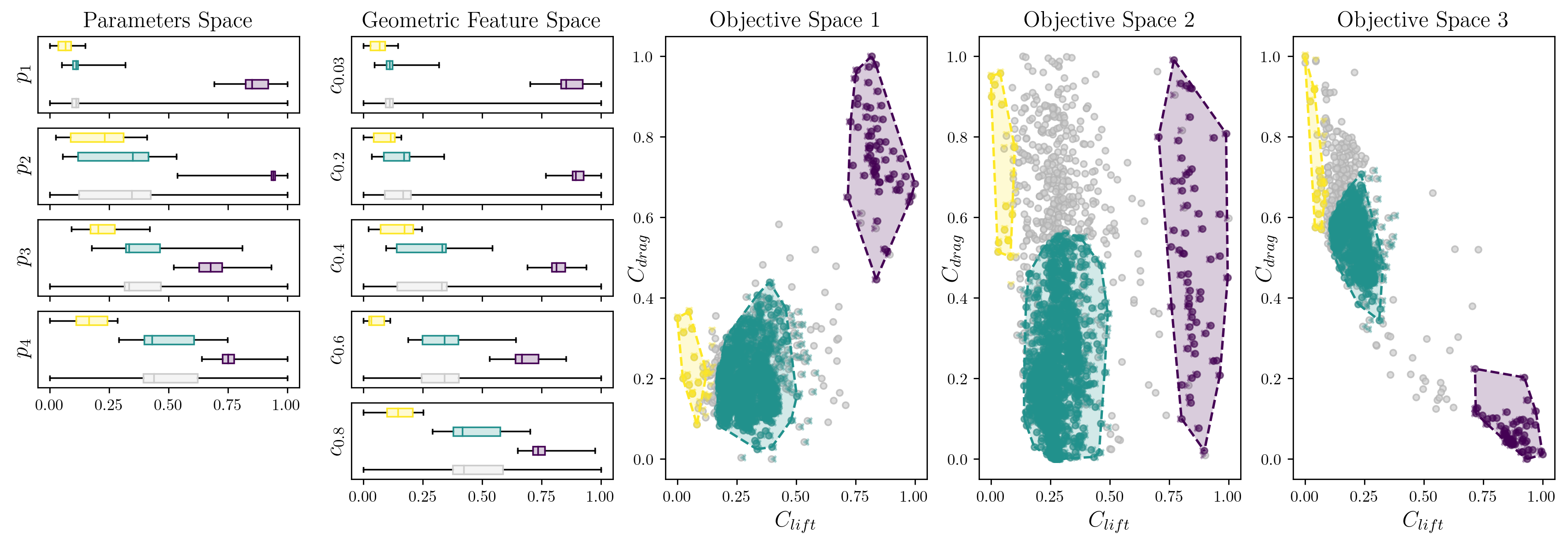}
\caption{Result of the concept identification process in five description spaces. Three concepts (purple, green and yellow
samples) are identified that each cover parts of each description space. The colored regions indicate the concept candidate
regions $C_{\alpha,k}$. Inside each concept candidate region the samples actually belonging to concept $C_\alpha$ are also colored, while there
are still data samples in grey which are not associated with the corresponding concept} \label{fig:AF_Result}
\end{figure*}

\begin{figure}[!ht]
\centering
\includegraphics[width=0.3\textwidth]{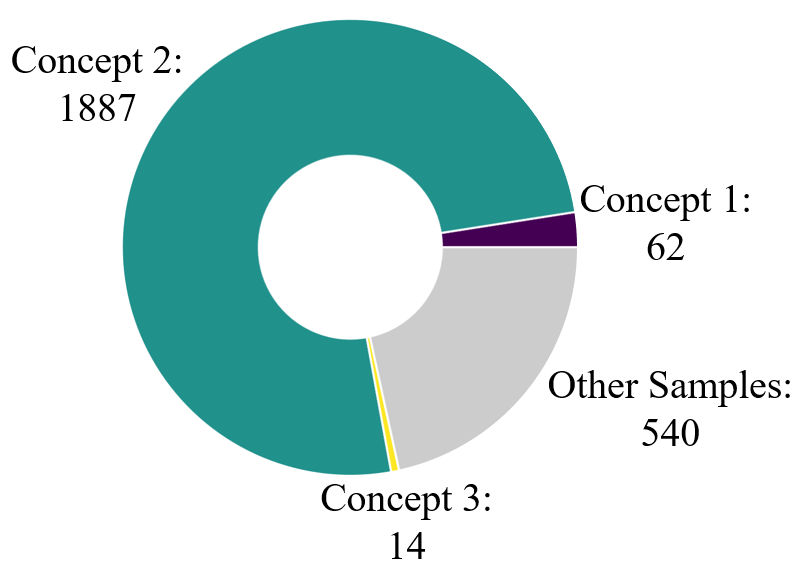}
\caption{Concept sizes.  Number of samples for each concept for the airfoil experiment with five description spaces.}
\label{fig:AF_Quality}
\end{figure}

The process again identifies three concepts which cover large portions of the data samples as shown in Figure~\ref{fig:AF_Result}.
Compared to the previous example with only four description spaces, the location of the concept is roughly similar.
However, the extent and size of each concept have changed significantly (Figure~\ref{fig:AF_Quality}). 
The number of samples not associated with any concept is strongly reduced and the green concept 2 now includes about 3/4 of all samples.
On the other hand, the yellow concept, previously the largest concept, is now the smallest concept and only comprises 14 samples. 
This is actually too small and consequently the CQM $Q$ incurs a size scaling factor which is slightly reduced from one, in particular $ F ( \tfrac{14}{2503},0.01)=0.9983$.
The CQM scores for each concept shown in Table~\ref{tab:metric_values} lead to a larger overall CQM of $Q=0.65$ than the previous example (where $Q=0.54$ was obtained), which is mainly due to the increased size and consistent definition of the green concept 2. 

Also in this example with five description spaces, the method is able to identify meaningful and consistent concepts which include user preference and respect size preferences.

\begin{table}[!ht]
\centering
\caption{Metric scores for the identified concepts in five description spaces. (*) The yellow concept is slightly too small which incurs an size-scaling factor of 0.9983. The total CQM for these concepts is $Q=0.65$.}
\begin{tabularx}{0.96\linewidth}{lcc}
\toprule
& \# Samples & CQM per concept $Q_\alpha$ \\
\midrule
\rowcolor{lightPurple}
Concept 1 & 62 & 0.97 \\
\rowcolor{lightGreen}
Concept 2 & 1887  & 0.95\\
\rowcolor{lightYellow}
Concept 3 & 14* & 0.71 \\
\bottomrule
\end{tabularx}
\label{tab:metric_values}
\end{table}
%

\subsection{Selection of Representatives}\label{subsec:selection}

One of the main use-cases for concepts is to select representative data samples for each concept and thereby mapping the design space.
For example, if the design shall be further optimized based on the discovered concepts, one option would be to select an archetype of each concept as an initial starting point for the additional optimization process.
A multitude of different methods for determining such archetypes from the concepts can be devised, ranging from random sampling designs from each concept to selecting Pareto-optimal subsets.
The concrete application and engineering problem to be solved determines the best suited method to be implemented.  

\begin{figure*}[!ht]
\centering
\includegraphics[width=0.96\textwidth]{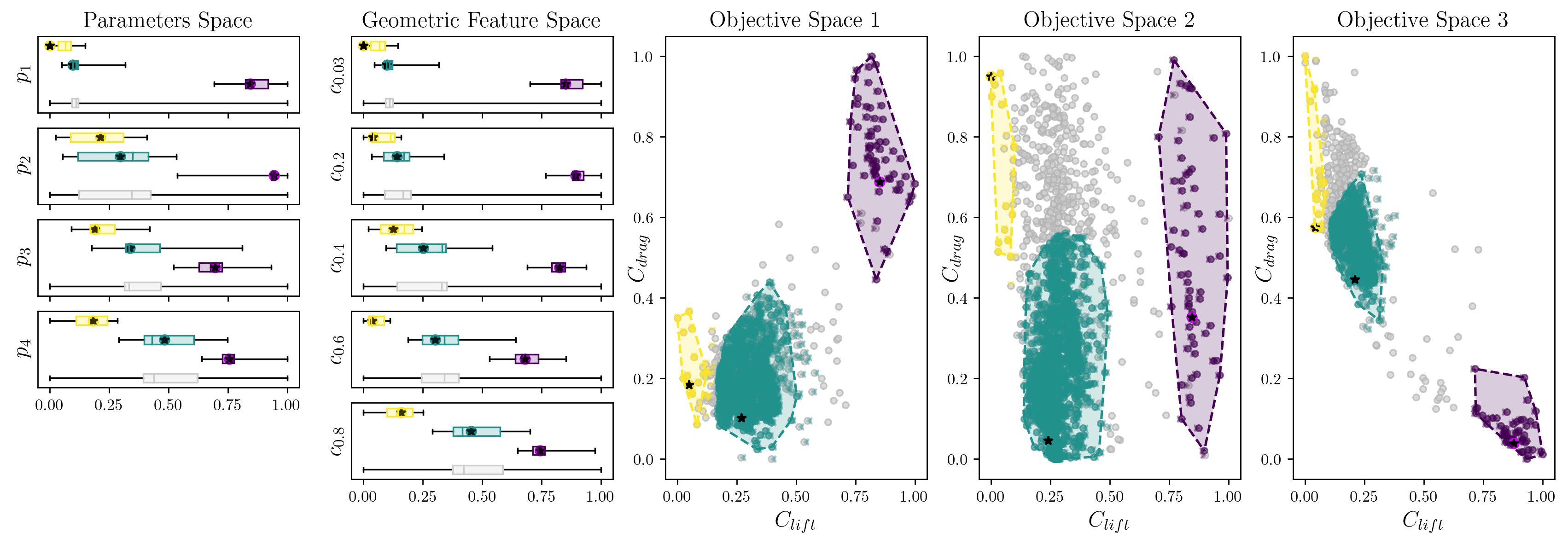}
\caption{Selection of concept-representatives. The samples marked with a star are the chosen representatives for each concept. The selection is based on proximity to the geometric mean of the respective concept in the parameter space.} \label{fig:AF_Selection1}
\end{figure*}

\begin{figure*}[!ht]
\centering
(a)\phantom{FFFFFFFFFFFFFFFFFFFFFFFFFFFFFFFFFFFFFFFFFFFFFFFFFFFFFFFFFFFFFFFFFFFFF}\\
\includegraphics[width=0.96\textwidth]{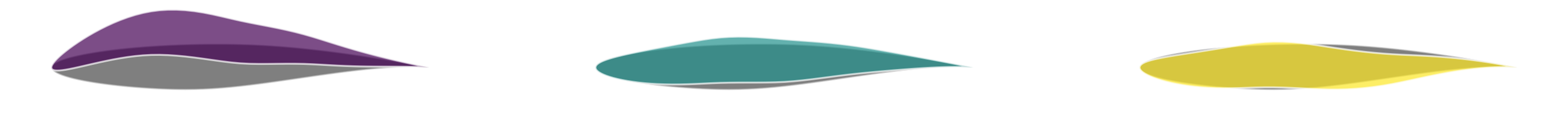}\\
(b)\phantom{FFFFFFFFFFFFFFFFFFFFFFFFFFFFFFFFFFFFFFFFFFFFFFFFFFFFFFFFFFFFFFFFFFFFF}\\
\includegraphics[width=0.96\textwidth]{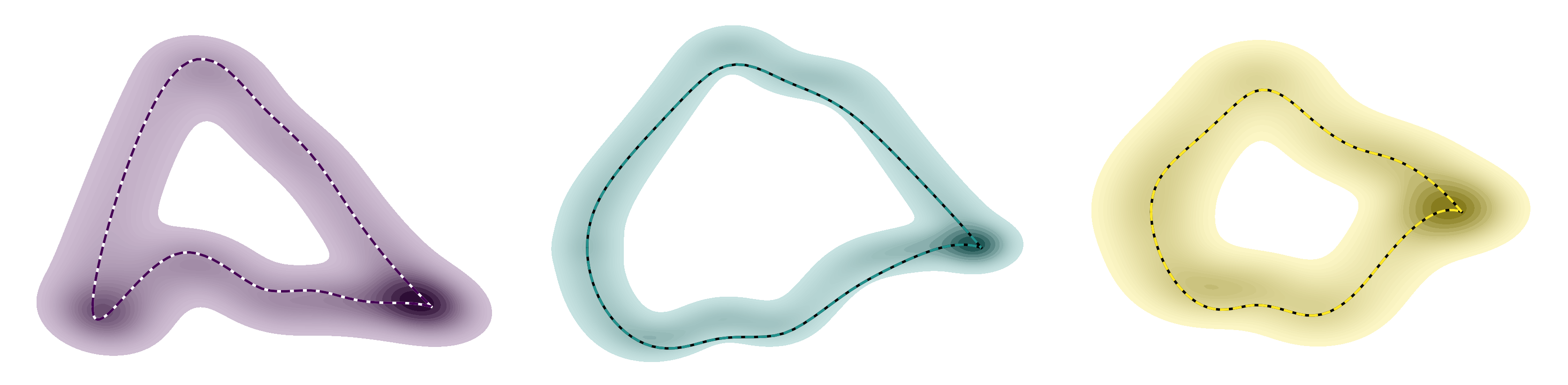}
\caption{Concept-representatives. (a) The colored concept representatives in comparison to the gray RAE2822 base profile. 
(b) Distribution of the concepts and their representatives 1: The kernel-density-estimation of the shapes of samples that are associated with each of the three concepts. The dashed lines show the selected representatives. The y-axis is stretched for better visibility.
} 
\label{fig:AF_Representatives_vs_Baseline}
\end{figure*}

As an example, the one sample for each concept, that is closest to the geometric mean of each concept in the parameter space is selected as archetype. 
Such a choice could be well motivated if the archetype should be used to initialize several refinement and optimization processes for each concept.  
The dataset with the identified concepts and the archetypes are depicted in Figure~\ref{fig:AF_Selection1}, while the archetypes themselves are shown in Figure~\ref{fig:AF_Representatives_vs_Baseline}. 
In Figure~\ref{fig:AF_Representatives_vs_Baseline}(b), the representative for each concept is marked along with an indication of the distribution of the other airfoils in that concept, which is obtained by performing a kernel-density estimation of the airfoil surfaces. 
It can be clearly seen that, for one the concepts share characteristic features of the airfoils, and for another that the archetypes represent these features.
It should be noted that while this analysis and the plots shown in Figure~\ref{fig:AF_Representatives_vs_Baseline} focus on the purely geometric aspects of the airfoils, the analysis and concept identification, which is the basis for this, does not.
Therefore, it can be expected that all samples from each concept also share similar performance values. 
This is not trivial, given the complex structure of the dataset discussed above. 

There is a multitude of other options to chose representatives from concepts, highly depending on the application and design task.
The presented example only demonstrates a simple example having subsequent optimization and redesign processes in mind.

\subsection{Discussion}\label{subsec:discussion}

The presented experiments demonstrate the general capabilities of the concept identification method based on optimizing the proposed CQM.
However, there are drawbacks. 
Similar to previously proposed methods~\cite{Lanfermann2020}, the process relies on an optimization algorithm to optimize the defined quality metrics. 
Depending on the dimension of the search space, the dataset, and the optimization algorithm, the optimization problem might be hard, rendering good solutions difficult to find. 
Also, performing multiple optimization runs typically results in different concepts, where the initialization of the search process has a significant influence. 
On one side, this might be seen as a drawback, as the proposed concept identification method does not find optimal concepts in a deterministic fashion.
On the other side, this can be viewed as an advantage, since the concept identification problem is not well defined and different users might prefer different concept definitions, also depending on their current application problem at hand. 
Therefore, a method like the propose one which can produce different but still good results, will provide more options to the engineer.

\begin{figure*}[!ht]
\centering
\includegraphics[width=0.96\textwidth]{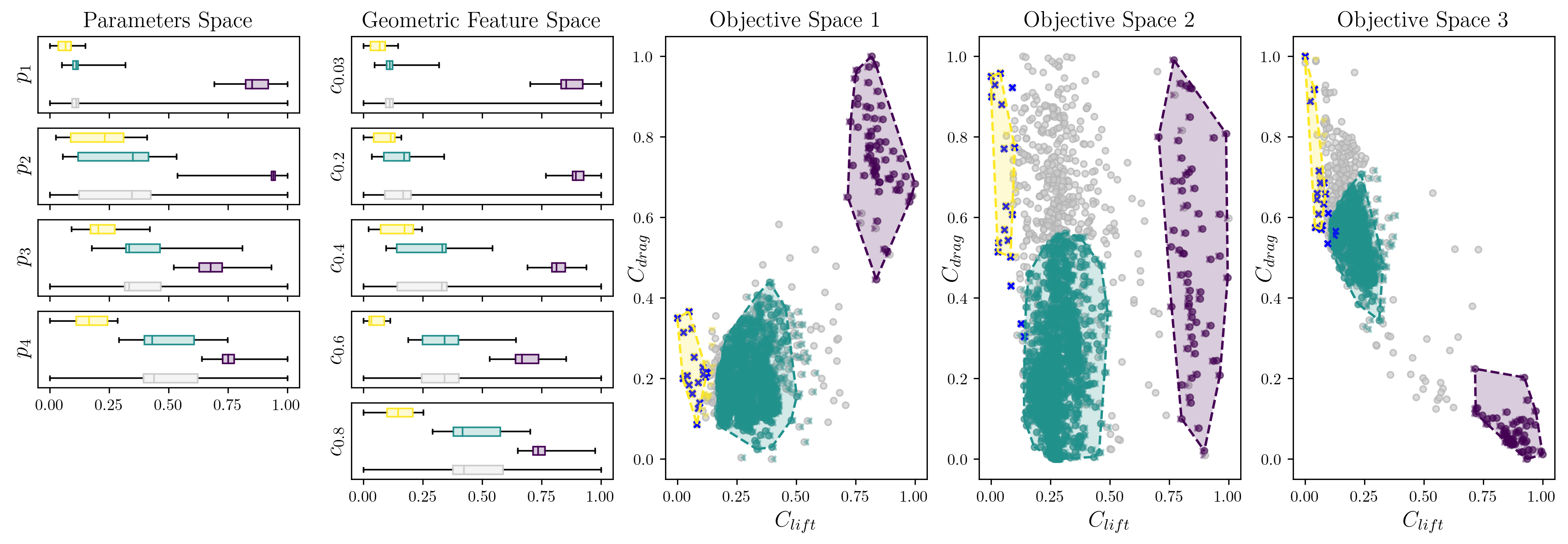}
\caption{Influence of the parametrization of the concept candidate regions. The blue crosses mark all samples that are encompassed by the yellow concept in objective space 1. In the other objective spaces, they cannot be captured with the current ellipse-shaped concept candidate regions without also integrating many other samples into the concept.} \label{fig:AF_Discussion}
\end{figure*}

Another issue of the current implementation is the necessity to choose and retain a specific parametrization of simply connected regions which are suited to select candidate samples for each concept.
In the experiments, hyper-ellipsoids were used which proved a reasonable tradeoff between shape-variability and number of search parameters. 
However, this choice obviously restricts the possible shape of the concepts during the optimization.
Non-convex concept sets or thin asymmetrical shapes cannot be represented with ellipsoids. 
Due to this, some direct improvement on the concepts of the above examples cannot be realized with such a parametrization. 
For example, all samples that lie within the convex hull of the yellow concept 3 are shown in blue in objective space 1 (Figure~\ref{fig:AF_Discussion}).
Some, but not all of these samples, belong to this concept as can be seen since they are outside the yellow region on other description spaces. 
However, it can be imagined that at least some of those points could be added to the yellow concept 3.
This could be done by simultaneously extending the region in objective space 2 and 3 by a very narrow and non-convex shape along the left edge of the dataset and slightly adjusting the region of the green concept 2. 
However, the parametrization of the shapes as ellipsoids  does not allow for this.

\section{Conclusion and Outlook}\label{sec:conclusion}

In this work, an approach to define meaningful and consistent concepts in engineering datasets was presented.
The notion of description spaces was introduced where the features associated with the designs in the dataset are partitioned into separate groups.
Each such group constitutes one description space.
At the core of the method is a novel concept quality  measure (CQM). 
It considers the number of samples associated with each concept while avoiding samples to be associated with multiple concepts, i.e.\ concept overlap. 
The metric also accounts for user preference.

The properties of the proposed CQM were thoroughly discussed and it was shown to be well suited to deal with the complex structure of engineering datasets incorporating a multitude of different features. 
In particular, it was highlighted that the CQM is able to handle the complex structure of typical engineering datasets, where some samples in a region associated with a concept cannot be associated with that concept due to, for example, incomplete feature sets.
The usefulness of the proposed approach was demonstrated for a realistic dataset derived from computational fluid dynamics simulations of airfoil profiles.
The identified concepts were used to select archetypal representatives for each concept, which were found to faithfully represent the geometries in the concepts with their characteristic features.

In future work, the current implementation can be improved by using alternative parameterizations of the region for selecting the candidates for each concept. 
This avoids the described shortcomings of the used hyper-ellipses and allows to form more refined concepts shapes and thus better concept definitions.
A technical drawback is given by the high computational demand of the method. 
The optimization problem employed for identifying the optimal concepts becomes increasingly harder with increasing number of features for each design. 
Therefore, it would be desirable to formulate a constructive approach based on the proposed CQM which can generate concepts with high quality directly.  

Even though natural partitions of the features typically exist, the definition of the description spaces is arbitrary. 
This consequently leads to the question, what the influence of different partitioning is on the resulting concepts, and if an optimal partitioning exist for a specific problem.
This is in particular relevant for applications with semantically more heterogeneous objectives, where a partitioning of the features into description spaces is not obvious. 
Therefore, thorough investigation of more diverse datasets will allow for a refinement of the concept identification approach and guidelines for the partitioning of features into description spaces should be proposed.

\section*{Acknowledgements}
The authors would like to thank Stefan Menzel for the fruitful discussions and constructive comments on the manuscript.   



\end{document}